\documentclass[letterpaper, 10 pt, conference]{ieeeconf}  

\usepackage{booktabs}
\usepackage{graphicx}
\usepackage{subcaption}

\IEEEoverridecommandlockouts                              

\usepackage{mathptmx} 
\usepackage{times} 
\usepackage{amsmath} 
\usepackage{amssymb}  

\title{\LARGE \bf
Motion-Focused Latent Action Enables Cross-Embodiment VLA Training from Human EgoVideos
}

\author{Runze Xu$^{1,2}$, Yiluo Zhang$^{2}$, Jian Wang$^{1}$, Yu Wang$^{1}$, Jincheng Yu$^{*1}$
\thanks{$^{1}$ Department of Electronic Engineering, Tsinghua University}
\thanks{$^{2}$ Tianfu Jiangxi Laboratory}
\thanks{$^*$ Corresponding at: yu-jc@mail.tsinghua.edu.cn}%
}

\begin{document}

\maketitle
\thispagestyle{empty}
\pagestyle{empty}

\begin{abstract}

Training generalist Vision-Language-Action (VLA) models typically requires massive, diverse robotic datasets with high-fidelity action annotations. While egocentric human manipulation videos are abundant and capture significant environmental diversity, the absence of action labels makes them difficult to use in conventional training paradigms. To address this, we propose a latent-action-based framework designed to extract general action priors from unlabeled human videos. The architecture features a Hybrid Disentangled VQ-VAE that decouples motion dynamics from environmental backgrounds through physical masks, enabling the construction of a cross-embodiment action codebook. By pre-training on human videos with the codebook, the VLM backbone learns deep representations of action intent. For adaptation to specific embodiments, we introduce an intent-perception decoupling strategy where the VLM predicts the action intent while a separate frozen visual encoder provides state-specific features to the action expert, thereby reducing action hallucinations. Results in simulation and real-world environments show that our method, pre-trained exclusively on unlabeled human videos, performs competitively with state-of-the-art VLA models trained on massive annotated datasets, requiring only $\sim 50$ trajectories for downstream adaptation. Code is available at https://github.com/shuiyigong/Motion-Focused-Latent-Action.

\end{abstract}

\begin{figure*}[ht]
  \centering
  \includegraphics[width=\linewidth]{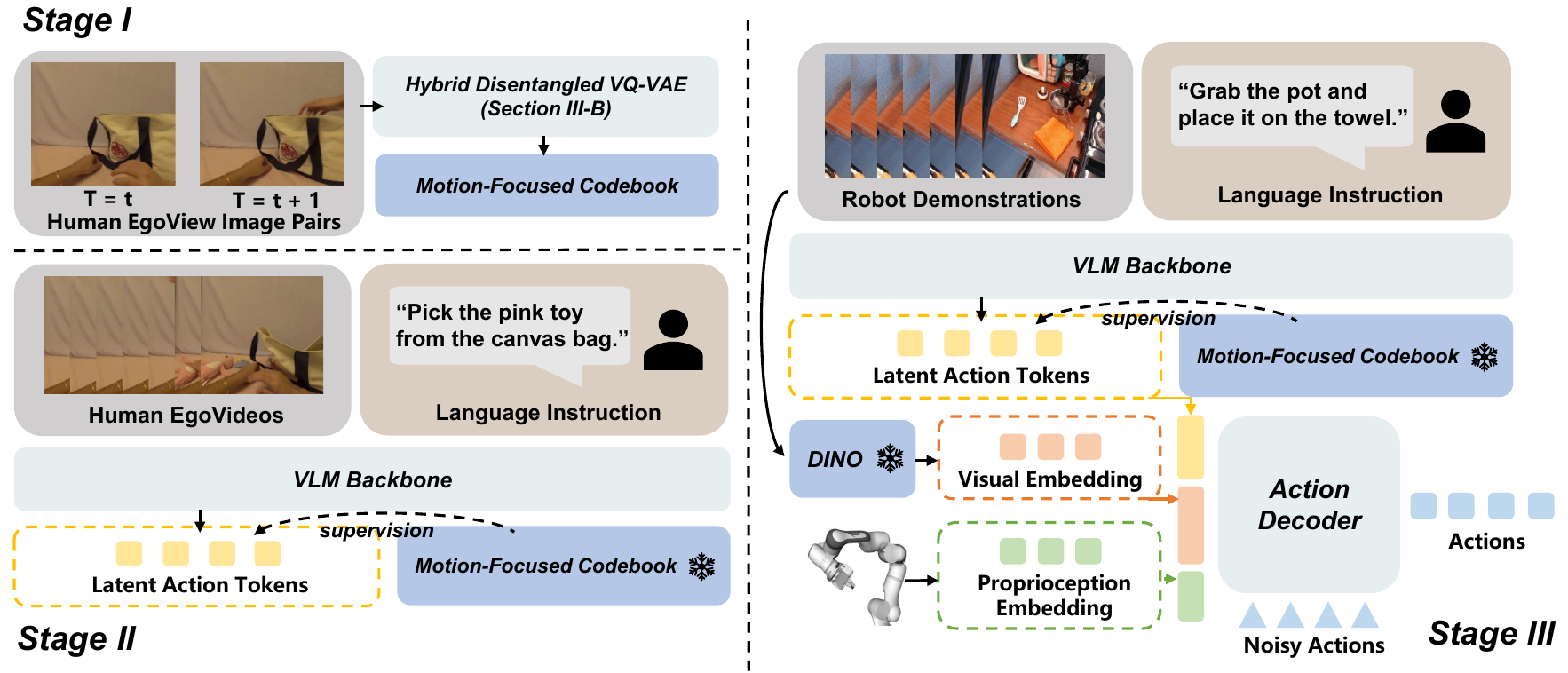}
  \caption{\textbf{Method overview.} We propose a human-video-driven framework for training vision--language--action models.
A hybrid disentangled VQ-VAE first extracts transferable latent action codes from unlabeled human videos. These codes are then used as supervision to pre-train the VLM to infer action intentions from observations and instructions. Finally, with only a small number of robot trajectories, the VLM backbone and action expert are jointly fine-tuned to ground the learned intentions in real robotic execution.}
  \label{fig:frame}
\end{figure*}
\section{INTRODUCTION}

Recent progress in Vision-Language-Action (VLA) models, ranging from early architectures like RT-2\cite{zitkovich2023rt} to recent systems such as pi0\cite{black2024pi_0} and RDT\cite{liu2024rdt}, has demonstrated the efficacy of using powerful Vision-Language Models (VLMs) for cross-scenario execution. These models achieve high performance by fine-tuning on downstream task datasets. However, training general VLA models remains heavily dependent on large-scale robotic datasets with precise action annotations, such as Open X-Embodiment\cite{o2024open} and AgiBot\cite{bu2025agibot}. Collecting such data is not only prohibitively expensive in terms of time and equipment but also introduces a significant domain gap due to the kinematic and physical differences between robot platforms. This gap necessitates complex alignment algorithms to manage noise and inconsistencies across embodiments.

Egocentric human manipulation videos offer a scalable alternative because they are abundant on the internet and capture high environmental diversity. These videos reflect real-world complexities that far exceed those of laboratory or factory settings. Although recent studies such as EgoMimic\cite{kareer2025egomimic} and MotionTrans\cite{yuan2025motiontrans} attempt to utilize human data for post-training, and H-RDT\cite{bi2025h} explores pre-training with human videos, these works still rely on explicit hand-motion labels captured by \textbf{specialized and expensive AR or VR hardware setups}. Since the vast majority of internet-scale videos lack such annotations, current VLA models cannot directly leverage the most extensive human visual priors, leaving most human data untapped.

Beyond the labeling bottleneck, a fundamental difficulty in utilizing human videos lies in the entanglement of action representations. Recent approaches\cite{ye2024latent, chen2024igor} typically learn discrete action codebooks via VQ-VAE-style reconstruction objectives, compressing frame transitions into latent action tokens. However, these objectives often capture task-irrelevant dynamics, such as background changes or camera shifts. Such \textbf{noisy representations} introduce distractions unrelated to the actual manipulation tasks, which degrades the quality of pre-trained policies\cite{zhang2025latentactionmodelsactually}. UniVLA\cite{bu2025univla} attempts to guide the latent action model with linguistic descriptions to focus on task-centric movements. However, the vast complexity and environmental diversity of human recordings make it difficult to achieve precise decoupling through language signals alone, leaving significant room for improvement in human-to-robot transfer performance.

To address these issues, we propose a latent action framework that extracts universal action priors directly from unlabeled egocentric videos. We treat human videos as natural carriers of action intent rather than mere visual backgrounds. By pre-training VLMs with a discrete action codebook that captures embodiment-agnostic motion primitives disentangled from specific robot kinematics, the VLA model can acquire an understanding of action intent before exposure to actual robotic environments. To ensure stable execution, we also introduce an intention-perception decoupling strategy during adaptation that balances high-level VLM intent with objective physical feedback. Consequently, the downstream adaptation phase requires only few robotic trajectories to jointly train the VLM and the action head, achieving performance comparable to state-of-the-art VLA models across different embodiments.

This work makes several primary contributions.
\begin{itemize}
  \item First, we establish a pre-training paradigm that leverages unlabeled human videos to learn cross-embodiment action priors, enabling a transition from human action intent to robot execution with only $\sim 50$ downstream trajectories. 
  \item Second, we introduce a hybrid disentangled VQ-VAE that separates motion dynamics from environmental backgrounds via physical masks, ensuring the latent action space focuses on pure motion patterns. 
  \item  Finally, we validate the framework across cross-embodiment scenarios, including robot-to-robot and human-to-robot transfers. Experiments show the learned representations maintain high cross-embodiment consistency.
\end{itemize}

\section{RELATED WORK}
\subsection{Vision-Language-Action models}
Current Vision-Language-Action (VLA) models typically combine a pre-trained VLM backbone with a dedicated action head\cite{kim24openvla,liu2024rdt,black2024pi_0,nvidia2025gr00tn1openfoundation}. In this architecture, the pretrained VLM interprets instructions and visual observations into semantic embeddings, which then condition the action head to generate control sequences. For action head part, early implementations like OpenVLA\cite{kim24openvla} utilized autoregressive behavior cloning, and more recent developments\cite{liu2024rdt,black2024pi_0} have shifted toward generative policies, such as diffusion\cite{chi2025diffusion} or flow matching\cite{lipman2023flowmatchinggenerativemodeling}, to better represent the multimodal nature of robot actions. However, these systems remain heavily dependent on large-scale, annotated robot trajectories for pre-training. Such data is inherently difficult to collect and standardize across different robotic embodiments compared to internet-scale text or images, creating a bottleneck for the scalability of VLA systems.

\subsection{Latent Action Representation Learning}
Latent action representation learning extracts generalizable control signals from unlabeled video data. Early frameworks, such as LAPA\cite{ye2024latent} and IGOR\cite{chen2024igor}, utilize VQ-VAE architectures to compress features from consecutive frames into discrete latent codes. By reconstructing subsequent frames conditioned on these latents, these models learn codebooks that represent essential visual transitions. UniVLA\cite{bu2025univla} builds on this foundation by introducing a two-stage training strategy designed to suppress task-irrelevant variations within frame transitions. Further developments, including villa-x\cite{chen2025villa}, integrate robot-specific states and actions into the self-supervised process to ground latent representations in physical motion patterns, while GO-1\cite{bu2025agibot} and GR00T1\cite{nvidia2025gr00tn1openfoundation} incorporate explicit supervision over these latent spaces during training. A major difficulty remains in isolating motion from complex backgrounds in human videos without robot-side labels. We address this by explicitly decoupling motion from environmental noise using physical masks in our hybrid disentangled VQ-VAE, which distills pure action intent to improve cross-embodiment training.

\subsection{Using Human Data in VLA Training}
Early efforts to incorporate human data, such as UMI\cite{chi2024universal}, use handheld grippers to collect data, while subsequent works such as EgoMimic\cite{kareer2025egomimic} and MotionTrans\cite{yuan2025motiontrans} employ AR/VR hardware to extract hand poses as explicit action labels. Although H-RDT\cite{bi2025h} extends this by pretraining VLA on human video, it still relies on pre-annotated action sequences from VR devices, limiting their scalability similarly to robot-specific datasets. Furthermore, frameworks like UniVLA still require robot-side trajectories to ground their codebooks, which can lead to weak cross-embodiment generalization and diminished effectiveness in complex bimanual tasks. Our work bypasses the need for such annotations by learning action priors directly from unlabeled human videos.

\begin{figure}[ht]
  \centering
  \includegraphics[width=\linewidth]{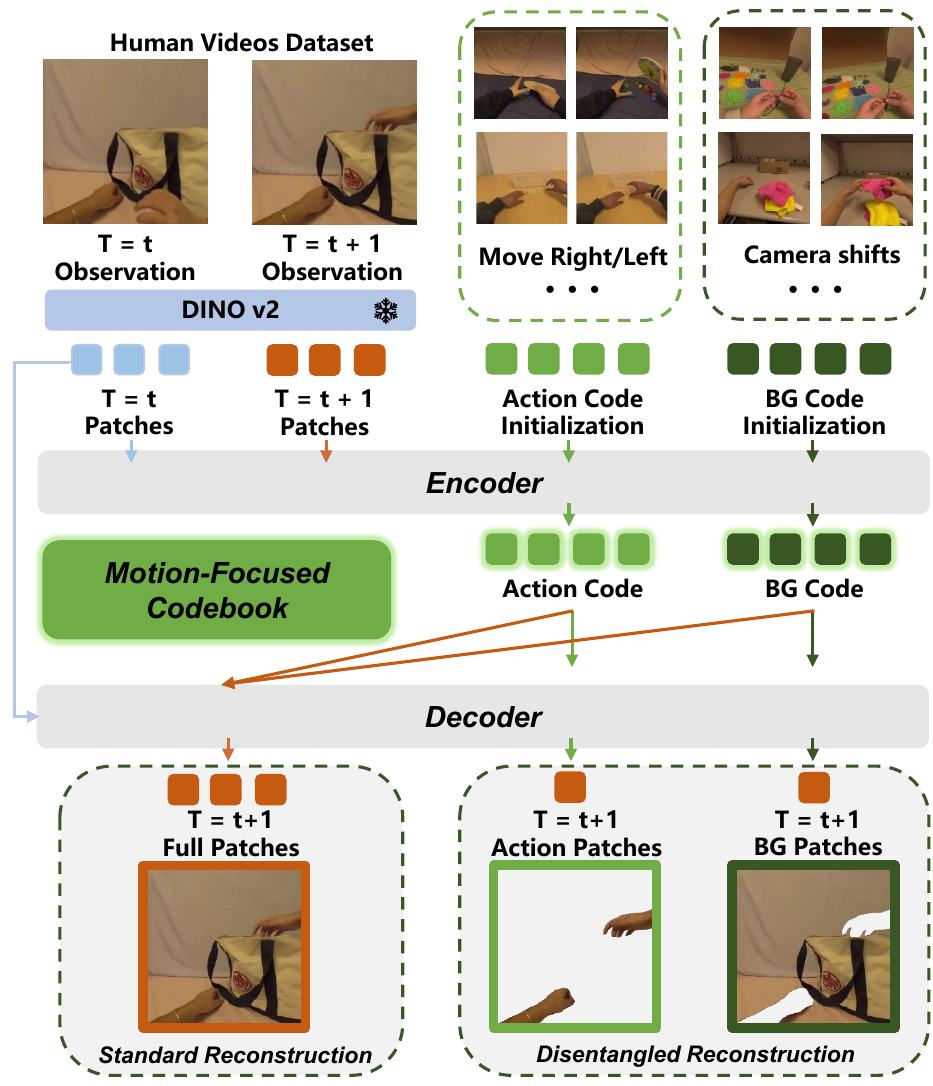}
  \caption{\textbf{Hybrid Disentangled VQ-VAE.} The VQ-VAE model decomposes short-term visual changes into discrete action and background latent spaces via a dual-path vector quantization bottleneck. A shared mask-guided decoder enforces semantic separation by reconstructing motion-related and background regions from corresponding latent codes, enabling the extraction of transferable action intentions from videos.}
  \label{fig:vae}
\end{figure}

\section{METHOD}

\subsection{Overview}
As shown in Fig.~\ref{fig:frame}, our framework learns robot manipulation policies from unlabeled human videos through three main stages. In the first stage, we design a hybrid disentangled VQ-VAE to extract cross-embodiment discrete action codebooks from egocentric human datasets. In the second stage, these codes provide the supervision for VLM pre-training, where a vision-language model learns to map visual observations and language instructions to general action intentions. Finally, in the downstream adaptation stage, the model is fine-tuned on a target robot platform using a limited set of demonstrations. This stage integrates the distilled action intent with visual feedback from a DINO v2 encoder to generate control commands via flow matching.

\subsection{Hybrid Disentangled VQ-VAE}

The latent action model uses a hybrid disentangled VQ-VAE to separate the latent spaces of action intent and environmental background changes. As shown in Fig.~\ref{fig:vae}, the architecture consists of a disentangled encoder, a dual-channel quantization bottleneck and a mask-guided decoder. Given a sequence of adjacent video frames $V \in \mathbb{R}^{T \times C \times H \times W}$, the model learns two independent discrete latent spaces: an action space $\mathbf{Z}_{act}$ and a background space $\mathbf{Z}_{bg}$. This decomposition allows the model to partition complex temporal visual variations into executable action intent and environmental representations. In our implementation, we use pairs of adjacent frames with a fixed 1-second interval, which allows the model to capture short-term visual changes from physical actions while ignoring long-term scene drift.

\subsubsection{Encoder}

To extract features that are semantically consistent yet sensitive to geometric variations, a frozen DINO v2 is used to extract high-dimensional spatial representations $F \in \mathbb{R}^{T \times N \times D}$. A spatial-temporal transformer models the evolution of these features across consecutive frames. To ensure disentanglement and mitigate interference from camera motion or background noise, the architecture introduces two sets of learnable query latents, $\mathbf{Q}_{act}$ and $\mathbf{Q}_{bg}$. These queries are concatenated with visual patches and processed by the encoder, which then projects the output into separate latent representations for action and background information.

\subsubsection{Dual-path Vector Quantization and Disentangled Bottleneck}

The bottleneck contains two independent vector quantization layers for action and background information. In the action branch, the continuous features corresponding to action queries are mapped into a discrete latent space using an action codebook of size 16, producing a set of quantized latent codes 
$\mathbf{z}_q^{act}$. Each pair of frames is represented by 4 discrete latent action tokens, which together encode the visual changes from manipulator interaction. Simultaneously, a background codebook of size 16 maps features from background queries to $\mathbf{z}_q^{bg}$, capturing environmental information such as scene layout. This dual-path architecture enforces semantic isolation by constraining different feature types into predefined discrete slots.

\subsubsection{Mask-guided Reconstruction with Shared Decoder}

A shared spatial-temporal transformer decoder uses ablation-based reconstruction with three input combinations to maintain latent space purity. The full reconstruction path combines $\mathbf{z}_{q}^{act}$, $\mathbf{z}_{q}^{bg}$, and initial frame features to reconstruct the target feature map, which is standard practice in latent action model training. The action ablation path utilizes only $\mathbf{z}_{q}^{act}$ and the initial frame features to force the model to reconstruct regions specifically affected by motion. The background ablation path uses only $\mathbf{z}_{q}^{bg}$ to focus on environmental reconstruction. During training, external masks provide inductive biases: the action path calculates reconstruction error only for foreground regions, such as the robot arm, while the background path is supervised by the background regions. This asymmetric strategy compels the decoder to reconstruct specific regions based on the latent code type within a shared parameter space. In the implementation stage, We use SAM2\cite{kirillov2023segany} to segment the human hand and generate physical foreground masks for the action reconstruction branch.

\subsubsection{Optimization Objectives}
The total loss function is a weighted sum of reconstruction, vector quantization, and commitment components:
\begin{equation}
L_{\text{total}} = \lambda_{\text{recon}} L_{\text{recon}} 
                   + \lambda_{\text{vq}} L_{\text{vq}} 
                   + \lambda_{\text{commit}} L_{\text{commit}}.
\end{equation}
The reconstruction loss $L_{recon}$ includes mask-guided foreground, background, and global feature errors. The vector quantization loss $L_{vq}$ optimizes codebook vectors by minimizing the Euclidean distance between encoder outputs and codebook entries, while the commitment loss $L_{commit}$ enhances training stability by preventing frequent fluctuations in encoder outputs. End-to-end optimization enables the unsupervised extraction of disentangled latent action representations.

\subsection{Pre-training VLM}
Since human video data lacks explicit action labels, joint training with a downstream action expert is not possible during VLM pre-training. To address this, we integrate the discrete action codebook learned by the hybrid disentangled VQ-VAE into the VLM vocabulary following UniVLA. For a given image pair $(I_t, I_{t+T})$ and its corresponding language instruction $L$, we first obtain the target latent action sequence $\mathbf{z}_{act} = \{z^{(1)}, z^{(2)}, \dots, z^{(K)}\}$ via the frozen VQ-VAE encoder, where $K=4$ in our implementation. The VLM is trained to predict this sequence auto-regressively. The pre-training objective is to minimize the negative log-likelihood:
\begin{equation}
L_{pre} = -\mathbb{E}_{(L, I_t, I_{t+T}) \sim \mathcal{D}} \left[ \sum_{k=1}^{K} \log P_\theta \left( z^{(k)} \mid z^{(<k)}, I_t, L \right) \right],
\end{equation}
where $z^{(k)}$ denotes the $k$-th action token, $z^{(<k)}$ represents the preceding action tokens in the sequence, and $\theta$ denotes the trainable parameters of the VLM. By distilling the VQ-VAE's disentangled latent space into the VLM, the model learns to ground high-level linguistic commands into low-level executable action intentions. In our implementation, we employ Prismatic-7B~\cite{karamcheti2024prismatic} as the backbone, ensuring a fair comparison with related baseline methods.

\subsection{Downstream Adaptation}
The adaptation stage transfers action intentions from the pre-trained VLM to specific robotic tasks. We use a transformer-based flow matching decoder as the action expert, which is trained from scratch to map latent intentions to executable control sequences. Given a natural language instruction $L$ and a primary visual observation $I_{main}$, the VLM generates a sequence of action tokens that represent the
predicted macro-level motion intent, and their corresponding hidden states from the last transformer layer are aggregated as the latent action embedding $f_{act}$. The VLM is updated via LoRA fine-tuning to achieve efficient specialization while preserving its general-purpose representations. The system constructs a multimodal context $F_{full}$ by concatenating the intent embedding with state features, including DINO v2 features $f_{obs} = DINO(I_{main})$ and robot proprioception $f_{proprio}$:
\begin{equation}
F_{full} = \text{Concat}(f_{act}, f_{obs}, f_{proprio}).
\label{eq:ffull}
\end{equation}
This context informs the transformer blocks of the action expert. Within these blocks, self-attention layers capture temporal dependencies within action chunks, while cross-attention layers integrate the multimodal context into the action sequence prediction. The model generates actions by predicting a vector field $v_{\theta}$ through an MLP head, with the flow matching loss:
\begin{equation}
L_{flow} = \| v_{\theta}(x_t, t, F_{full}) - (a - \epsilon) \|^2,
\end{equation}
where $a$ denotes the ground-truth action, $\epsilon$ is standard Gaussian noise, $x_t$ is the noisy action at flow step $t$, and $v_\theta$ is the velocity field predicted by the network. To ensure the VLM accurately captures the specific intention of downstream demonstrations, we also apply a cross-entropy loss $L_{intent}$ similar to loss in VLM pretraining stage to supervise the latent action token prediction. The final optimization objective is a joint loss that balances the intention prediction and the action execution:
\begin{equation}
L_{total} = L_{flow} + \lambda_{intent} L_{intent},
\end{equation}
where $\lambda_{intent}$ is a weighting hyperparameter. 

We use DINO v2 features as the visual representation instead of VLM visual embeddings to decouple action intent from the observed physical state. Directly using VLM embeddings for control often leads to action hallucination, where the model ignores real-time feedback. For instance, a model might attempt to place an object without perceiving that the target container remains closed. Separating the VLM-generated intent from the objective perception provided by DINO v2 mitigates these failures and ensures more grounded execution.

\section{Experiments}
This section systematically evaluates the proposed framework, focusing on whether the motion-focused latent action achieves robust generalization in cross-embodiment scenarios. The experimental design is structured around three core dimensions. First, in a robot-to-robot setting, we verify whether action intentions learned solely from single-arm videos generalize to different embodiments. Second, in a human-to-robot setting, we examine whether a dual-arm robot can be controlled using representations learned exclusively from ego-centric human videos without action labels. Third, we quantitatively analyze the representation space to determine if the learned latent action codebook maintains a consistent structure across different embodiments, thereby explaining the source of its cross-embodiment generalization capabilities. For all experiments, input images are resized to $224 \times 224$ pixels, and the DINOv2-ViT-B/14-reg model is utilized as the frozen visual encoder.

\begin{table}[t]
\centering
\caption{\textbf{Task success rates on the LIBERO simulation benchmark (\%).} Methods marked with * utilize wrist-mounted camera observations when post training.}
\label{tab:libero}
\renewcommand{\arraystretch}{1.15}
\resizebox{\linewidth}{!}{
\begin{tabular}{lccccc}
\toprule
\textbf{Method} 
& \textbf{Spatial} 
& \textbf{Object} 
& \textbf{Goal} 
& \textbf{Long} 
& \textbf{Average} \\
\midrule
LAPA\cite{ye2024latent}              & 73.8 & 74.6 & 58.8 & 55.4 & 65.7 \\
Diffusion Policy\cite{chi2025diffusion}  & 78.3 & 92.5 & 68.3 & 50.5 & 72.4 \\
OpenVLA\cite{kim24openvla}           & 84.7 & 88.4 & 79.2 & 53.7 & 76.5 \\
SpatialVLA\cite{qu2025spatialvla}        & 88.2 & 89.9 & 78.6 & 55.5 & 78.1 \\
pi0\cite{black2024pi_0}*               & 88.0 & 88.5 & 87.0 & 61.0 & 81.1 \\
pi0-fast\cite{pertsch2025fast}*          & 96.4 & 96.8 & 88.6 & 60.2 & 85.5 \\
villa-x\cite{chen2025villa}*           & 97.5 & 97.0 & 91.5 & 74.5 & 90.1 \\
UniVLA-Bridge\cite{bu2025univla}     & 90.0 & 86.0 & 90.5 & 86.0 & 88.1 \\
\midrule
Ours w/o DINO     & 91.0 & 81.0 & 88.5 & 81.0 & 85.4 \\
\textbf{Ours}     & 95.5 & 94.0 & 93.5 & 84.0 & \textbf{91.8} \\
\bottomrule
\end{tabular}
}
\end{table}

\begin{table*}[ht]
\centering
\caption{\textbf{Task success rates on the RoboTwin 2.0 simulation benchmark (\%).} Despite relying only on unlabeled human videos for pre-training, our method achieves performance comparable to state-of-the-art VLAs trained on large-scale robot datasets with action supervision. Ablation studies analyze the effects of the intention-perception decoupling strategy and examine the importance of latent action embeddings by freezing the VLM during post-training. \textbf{Bold} indicates the best performance, and \underline{underline} indicates the second best.}
\label{tab:robotwin}
\small
\renewcommand{\arraystretch}{1.15}
\resizebox{\textwidth}{!}{
\begin{tabular}{lcccccccc}
\toprule
\textbf{Task} 
& \textbf{RDT\cite{liu2024rdt}} 
& \textbf{pi0\cite{black2024pi_0}} 
& \textbf{ACT\cite{zhao2023learning}} 
& \textbf{DP\cite{chi2025diffusion}} 
& \textbf{UniVLA\cite{bu2025univla}} 
& \textbf{Ours w/o DINO} 
& \textbf{Ours (Freeze)} 
& \textbf{Ours} \\
\midrule
Adjust bottle        & 81 & 90 & \textbf{97} & \textbf{97} & 87 & 87 & 83 & \textbf{97} \\
Grab roller          & 74 & \underline{96} & 94 & \textbf{98} & 80 & 87 & 50 & 90 \\
Place phone stand    & 15 & \underline{35} &  2 & 13 & \textbf{40} & 30 &  8 & 28 \\
Pick dual bottles    & 42 & 57 & 31 & 24 & \textbf{58} & \textbf{58} & 42 & 52 \\
Place empty cup      & 56 & 37 & \underline{61} & 37 & 46 & 51 & 48 & \textbf{63} \\
Move can pot         & 25 & 58 & 22 & 39 & \underline{60} & 52 & 52 & \textbf{65} \\
Handover mic         & 90 & \textbf{98} & 85 & 53 & 82 & 86 & 84 & \underline{92} \\
Open laptop          & 59 & \underline{85} & 56 & 49 & 81 & 81 & 74 & \textbf{87} \\

Place object basket  & \textbf{33} & 16 & 15 & 15 & 18 & 16 & 11 & \underline{25} \\
Place burger fries   & 50 & \underline{80} & 49 & 72 & \textbf{84} & \underline{80} & 72 & 78 \\

\midrule
\textbf{Average}     & 52.5 & \underline{65.2} & 51.2 & 49.7 & 63.6 & 62.8 & 52.4 & \textbf{67.7} \\
\bottomrule
\end{tabular}
}
\end{table*}
\subsection{Single-Arm Robot to Single-Arm Robot}
This experiment validates whether action intentions learned from third-view robot videos can be transferred to different robotic embodiments for complex tasks. In this setup, we train the hybrid disentangled VQ-VAE and the VLM backbone using only the third-view video data from the BridgeV2 dataset\cite{walke2023bridgedata}, which features a WindowX manipulator, without using any action labels. The complete VLA model only encounters the Franka manipulator from the LIBERO\cite{liu2023libero} dataset during the post-training phase for action execution adaptation. During the training of the hybrid disentangled VQ-VAE, we employ RoboEngine\cite{yuan2025roboengine}, a state-of-the-art robot segmentation model, to generate high-quality physical masks that guide the decoupling of motion from the background.

The evaluation is conducted on the LIBERO benchmark, which is designed for lifelong robot learning and focuses on knowledge transfer across multiple tasks. The benchmark consists of four task suites, where LIBERO-Spatial evaluates adaptation to layout variations with similar semantic tasks, LIBERO-Object tests the transfer of manipulation skills to unseen objects, LIBERO-Goal assesses the understanding of diverse natural language instructions, and LIBERO-Long targets multi-stage manipulation requiring extended temporal dependencies. Each suite contains 10 tasks, and each task provides 50 human-teleoperated trajectories for post-training. We exclusively utilize the third-person view data from these 50 trajectories.

We compare our method against LAPA\cite{ye2024latent}, Diffusion Policy\cite{chi2025diffusion}, OpenVLA\cite{kim24openvla}, SpatialVLA\cite{qu2025spatialvla}, pi0\cite{black2024pi_0}, pi0-fast\cite{pertsch2025fast}, villa-x\cite{chen2025villa}, and UniVLA\cite{bu2025univla}. Among these, LAPA, villa-x, and UniVLA utilize latent action representation learning, while OpenVLA, SpatialVLA, and pi0 represent current mainstream VLA models. For pi0, we refer to the performance reported in villa-x. The codebook training and VLM pre-training phases for UniVLA are strictly reproduced on the BridgeV2 dataset. To ensure a fair comparison of latent actions, our method adopts the autoregressive action head used by UniVLA as the action expert. Furthermore, to verify the effectiveness of the decoupling between intention and perception, we substitute DINO v2 features for the visual embeddings output by the VLM. Post-training is conducted for 30k steps with a batch size of 128 for Spatial, Object, and Goal tasks, and 40k steps for Long tasks. Each task is evaluated 20 times.

Experimental results in Table~\ref{tab:libero} demonstrate that despite using only third-person videos and no action labels during pre-training, our method outperforms all baselines in the average success rate across the four suites.  While villa-x and pi0-fast maintain a slight lead in the Spatial and Object suites, their performance is bolstered by wrist-mounted camera observations and explicit action labels. Such egocentric feedback and direct supervision facilitate the fine-grained control required for short-range interactions. In contrast, our method excels in Goal and Long-horizon tasks, outperforming villa-x by 2.0\% and 9.5\% respectively. This performance gap indicates that our extracted latent action intentions provide superior guidance for multi-step sequences, effectively endowing the framework with stronger high-level planning capabilities. While villa-x excels in immediate motor precision, it lacks the global perspective necessary to maintain task coherence over extended periods. Our approach ensures that the robot remains focused on the final objective, mitigating the drifting issues common in long-horizon and goal-driven scenarios. Additionally, using DINO features instead of VLM visual embeddings yields superior results, confirming the efficacy of the intention-perception decoupling strategy in robot-to-robot generalization.

\begin{figure*}[ht]
  \centering
  \begin{subfigure}[t]{0.26\linewidth}
    \centering
    \includegraphics[width=\linewidth]{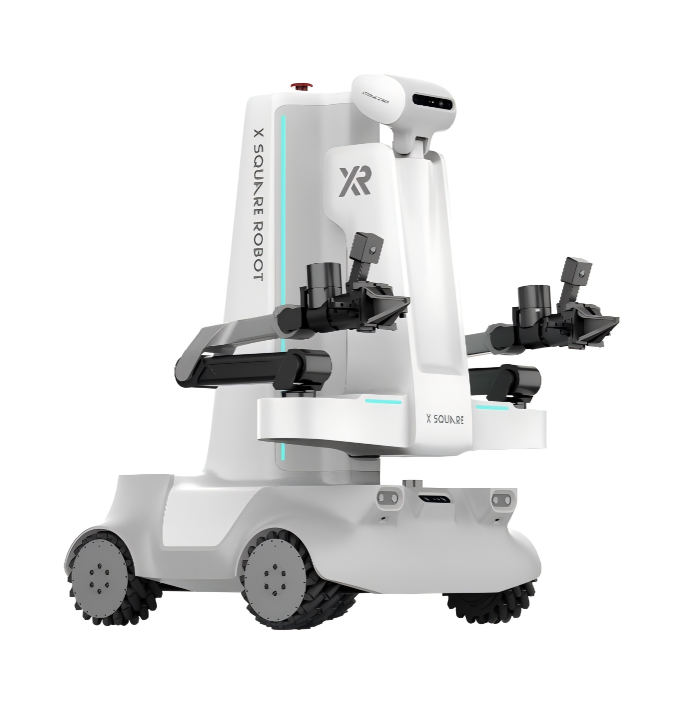}
    \caption{Robot platform}
    \label{fig:robot}
  \end{subfigure}
  \hfill
  \begin{subfigure}[t]{0.34\linewidth}
    \centering
    \includegraphics[width=\linewidth]{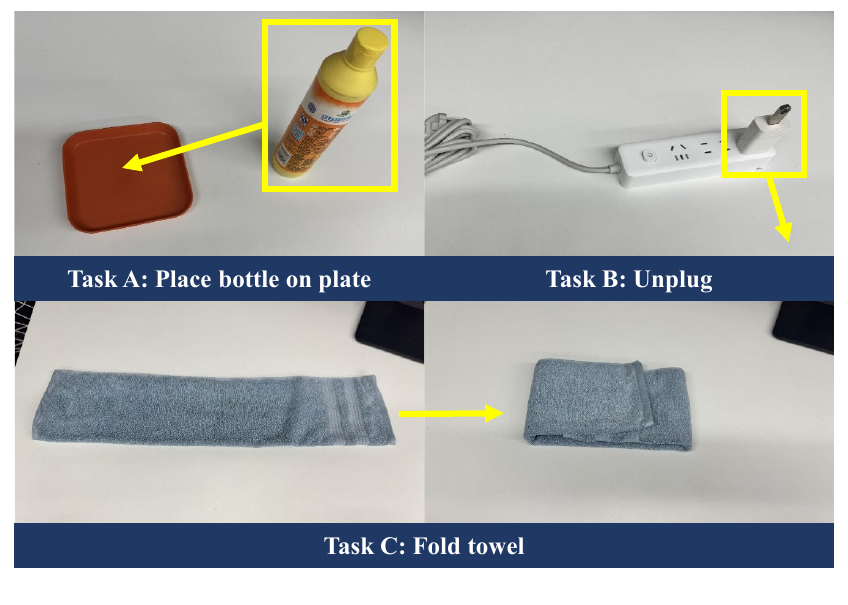}
    \caption{Three real-world tasks}
    \label{fig:realworld_task}
  \end{subfigure}
  \hfill
  \begin{subfigure}[t]{0.35\linewidth}
    \centering
    \includegraphics[width=\linewidth]{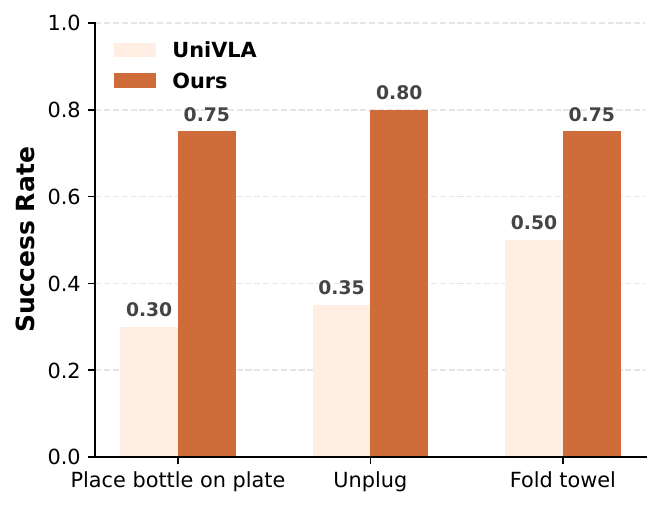}
    \caption{Success rates}
    \label{fig:realworld_results}
  \end{subfigure}

  \caption{\textbf{Real-world experiments.} (a) The physical dual-arm robot platform used for evaluation.
(b) Three real-world manipulation tasks, including placing a bottle on a plate, unplugging a power cord, and folding a towel.
(c) Task success rates compared with UniVLA, showing improved transfer of action intentions from human videos to the real robot. Notably, the ``Place Bottle” task shows a lower success rate. The bottle’s high center of gravity makes it prone to toppling, which results in irreversible failure within a trial, while the other tasks allow corrective retries.}
  \label{fig:realworld}
\end{figure*}
\subsection{Human to Dual-Arm Robot}
This section further investigates whether transferable action intentions can be learned from human ego-centric videos without action labels and generalized to dual-arm robotic manipulation.

We adopt RoboTwin 2.0\cite{chen2025robotwin} for simulation experiment. We train the hybrid disentangled VQ-VAE and VLM backbone using only the video portion of the EgoDex dataset\cite{egodex}. The VLA model only interacts with data collected from the RoboTwin 2.0 simulation environment during the post-training phase. During codebook training, SAM2 is utilized to generate physical masks for human hands to guide motion region modeling. The RoboTwin 2.0 benchmark includes fifty tasks, from which we select 10 representative tasks. We use the Aloha-Agilex dual-arm robot in clean mode to collect 50 trajectories per task for post-training, with a batch size of 32 and a total of 30k training steps for each task.

Due to frequent self-occlusion in the primary camera view caused by the robot arms in RoboTwin 2.0, we introduce an additional wrist-mounted view $I_{\text{wrist}}$ to enhance visual observability. Accordingly, the observation representation in Eq.~\eqref{eq:ffull} is reformulated as $f_{obs} = DINO([I_{main}, I_{wrist}]).$

Baselines include RDT\cite{liu2024rdt}, PI0\cite{black2024pi_0}, Diffusion Policy\cite{chi2025diffusion}, ACT\cite{zhao2023learning}, and UniVLA\cite{bu2025univla}. For UniVLA, a flow matching head is used as the action expert to ensure fairness. We also conduct two ablation studies, where one replaces DINO v2 features with VLM visual embeddings to verify the decoupling strategy, and the other freezes the VLM during post-training to examine the importance of latent action embeddings for downstream execution.

The results in Table~\ref{tab:robotwin} indicate that our method achieves performance comparable to state-of-the-art VLAs despite relying solely on unlabeled human videos for pre-training. The visual feature ablation proves that decoupling intention from state helps the model adjust action intentions based on real-time feedback. Freezing the VLM leads to a drop in success rates, demonstrating that the latent action embeddings output by the VLM are critical for downstream performance, as clear intentions guide the action expert. However, even with a frozen VLM, our model maintains a success rate close to RDT, which suggests that the VLM pre-training is highly effective. The action intentions extracted from human videos generalize directly to the robotic embodiment, as the latent action embeddings contain highly transferable information that guides control sequences even without further updates.

For real-world experiments, we verify the transferability of intentions to a physical dual-arm platform. The training setup mirrors the simulation, using only EgoDex for pre-training. The physical platform utilizes an ARX X5 leader arm to teleoperate an ARX R5 follower arm. Visual observations are provided by two wrist-mounted RGB fisheye cameras and one 20Hz first-person RGB camera. The model is deployed on the ARX R5 (Fig.~\ref{fig:robot}) at a 60Hz frequency. We evaluate three tasks (Fig.~\ref{fig:realworld_task}) : placing a bottle on a plate (single-arm), unplugging a power cord (dual-arm, rigid interaction), and folding a towel (dual-arm, deformable interaction). Each task uses 50 real-world trajectories for post-training with a batch size of 32 for 20k steps. Results in Fig.~\ref{fig:realworld_results} demonstrate that our method surpasses UniVLA in transferring intentions from human video data to the physical robot. Given the identical post-training data used across methods, these performance gains suggest that our framework extracts clearer action intentions, providing more explicit guidance for the action decoder and substantially improving downstream learning efficiency. Notably, UniVLA achieves a relatively low success rate on the ``Place Bottle” task due to the bottle’s high center of gravity, which makes it susceptible to irreversible toppling upon minor contact. Unlike the other two tasks, where failed attempts still allow for subsequent correction and retrying, errors in this task often result in terminal failure.
\begin{figure*}
  \centering
  \begin{subfigure}[t]{0.46\linewidth}
    \centering
    \includegraphics[width=\linewidth]{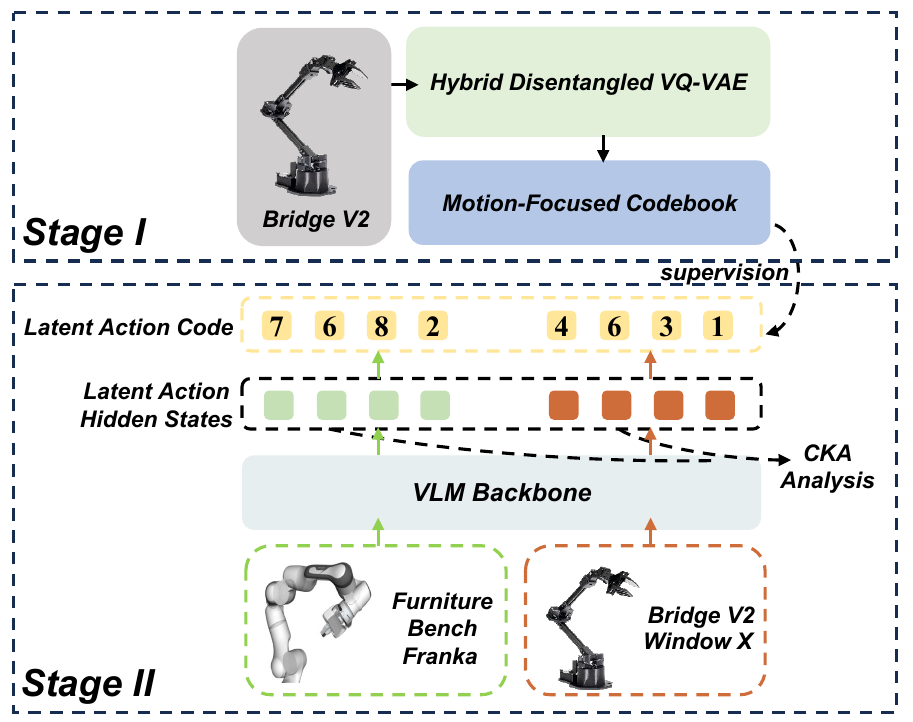}
    \caption{Latent Action Alignment Evaluation Overview}
    \label{fig:cka1}
  \end{subfigure}
  \hfill
  \begin{subfigure}[t]{0.52\linewidth}
    \centering
    \includegraphics[width=\linewidth]{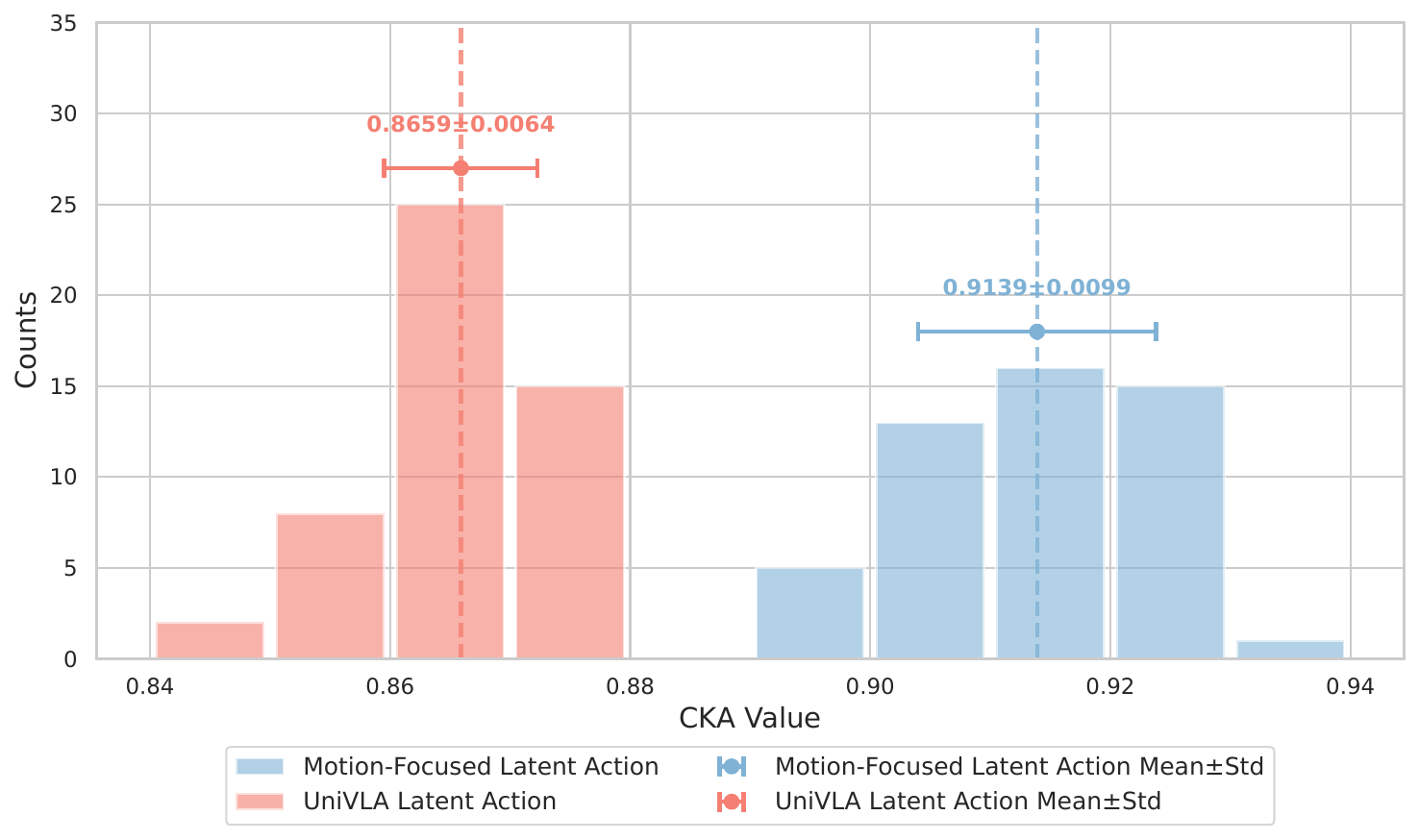}
    \caption{Latent Action CKA Analysis}
    \label{fig:cka}
  \end{subfigure}
  \caption{\textbf{Comparison of latent action alignment consistency.} The proposed Motion-Focused latent action outperforms UniVLA in CKA metrics, indicating a more coherent cross-embodiment action space.}
  \label{fig:ckaall}
\end{figure*}
\subsection{Latent Action Evaluation}
To explain the observed generalization performance at the representation level, we design an alignment analysis method based on domain subspace elimination. This approach quantitatively assesses the alignment of motion-focused latent actions in cross-embodiment scenarios and compares it with UniVLA. Specifically, as shown in Fig.~\ref{fig:cka1}, we first generate a discrete action codebook using a latent action model trained exclusively on BridgeV2 dataset. This codebook serves as pseudo-labels for pre-training the VLM on a mixed dataset of BridgeV2 (WindowX Robot) and FurnitureBench\cite{heo2023furniturebench} (Franka robot). We then extract feature embeddings corresponding to each action token from the hidden layers of different datasets for analysis. To ensure objectivity, 20,000 image pairs are randomly sampled from both datasets.

Since different environments and embodiments introduce domain bias, we employ an iterative domain subspace stripping strategy. A logistic regression classifier is trained to predict the domain source. If the accuracy exceeds chance, the features still contain domain-specific information. We then identify the principal component directions with the highest variance via PCA and treat them as the domain-related subspace. The original features are projected onto the orthogonal complement of this subspace to filter out domain-sensitive components. This process repeats until the classifier can no longer distinguish between data sources.

Following bias removal, we use the Centered Kernel Alignment (CKA) metric\cite{kornblith2019similarity} to compare latent action representations between BridgeV2 and FurnitureBench. For statistical robustness, we introduce a bootstrap sampling strategy based on token centroids. For each token appearing in both datasets, we collect its sample indices and, in each sampling round, randomly draw half of the samples to compute a temporary centroid. After 50 rounds, we calculate the CKA values between the resulting feature matrices.

The results (Fig.~\ref{fig:cka}) show that UniVLA exhibits lower consistency with a mean CKA of 0.8659, whereas our Motion-Focused Latent Action achieves a significantly higher alignment with a mean CKA of 0.9139. Both methods show minimal standard deviations, indicating robust results. This quantitative analysis confirms that our motion-focused latent action suppresses domain bias from environment and embodiment differences.

\subsection{Latent Action Analysis}
\begin{figure}
  \centering
  \includegraphics[width=\linewidth]{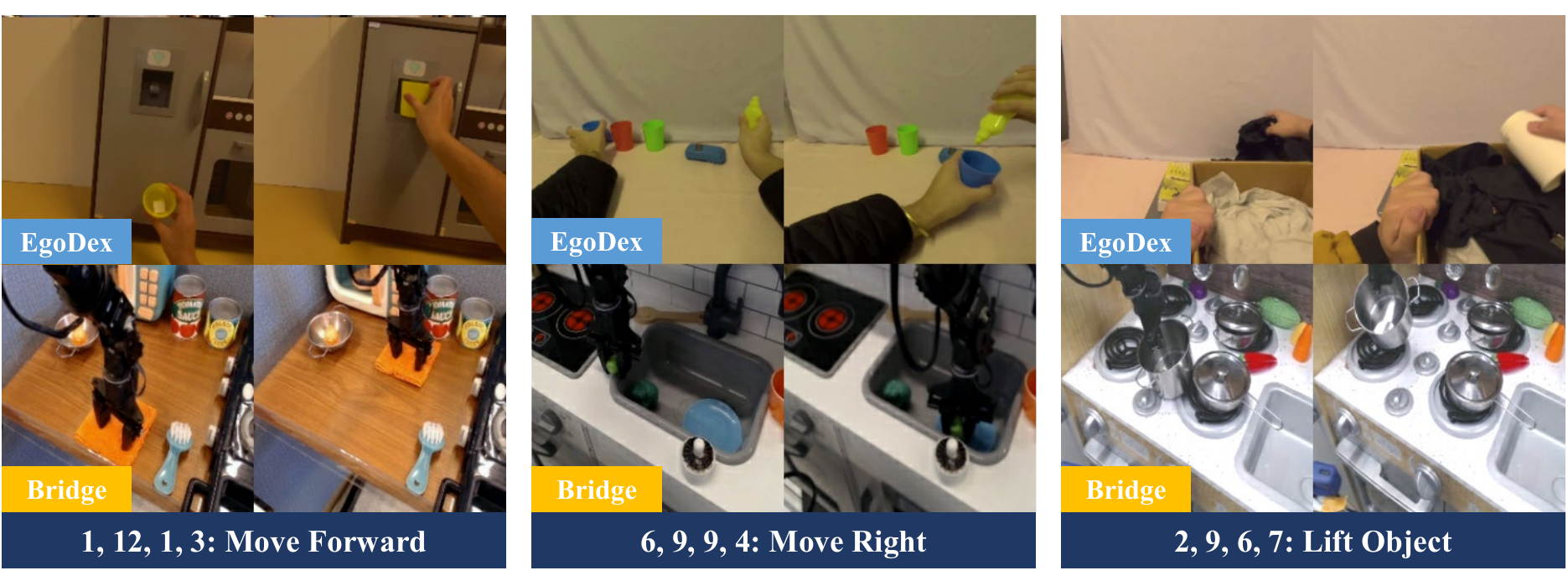}
  \caption{\textbf{Latent Action Visualization.} Image pairs from different datasets with same latent codes. Despite different morphologies, robot arms and human hands are assigned the same action tokens.}
  \label{fig:latentaction}
\end{figure}

Finally, to visualize cross-embodiment consistency, we train the hybrid disentangled VQ-VAE on a mixed BridgeV2 and EgoDex dataset and visualize the latent action codes for identical actions. As shown in Fig.~\ref{fig:latentaction}, robot arms and human hands are mapped to the same action tokens when performing the same behavioral patterns, further validating that the learned representations possess a highly consistent semantic structure across different embodiments.

\section{CONCLUSION AND LIMITATION}

In this paper, we introduce a latent action pre-training framework that connects unlabeled human video data with robotic control. By leveraging a hybrid disentangled VQ-VAE and an intention-perception decoupling strategy, our model achieves great performance with extremely high data efficiency. Despite these advancements, the proposed method has certain limitations. Although the discrete action codebook captures high-level intentions effectively, its representation capacity remains insufficient for fine-grained manipulation tasks requiring high-precision control. Future work will investigate multi-scale latent representations to improve the performance of VLA models in complex and delicate interaction scenarios.

\section*{ACKNOWLEDGEMENT}
This research was supported by National Natural Science Foundation of China (62325405), Tsinghua University Initiative Scientific Research Program, Tsinghua-Efort Joint Research Center for EAI Computation and Perception and SunRisingAI Lab, Beijing National Research Center for Information Science, Technology (BNRist), Beijing Innovation Center for Future Chips, and State Key laboratory of Space Network and Communications. 
\addtolength{\textheight}{-0.1cm}   




\bibliographystyle{IEEEtran}
\bibliography{IEEEabrv,refs}

\end{document}